%% file: main.tex
\pdfoutput=1
\documentclass[11pt]{article}

\usepackage[final]{acl}

\usepackage{times}
\usepackage{latexsym}

\usepackage[T1]{fontenc}
\usepackage[utf8]{inputenc}

\usepackage{microtype}

\usepackage{inconsolata}

\usepackage{bm}
\usepackage{xurl}
\usepackage{dsfont}
\usepackage{amsmath}
\usepackage{amssymb}
\usepackage{booktabs}
\usepackage{enumitem}
\usepackage{marvosym}
\usepackage{graphicx}
\usepackage{multirow}
\usepackage{tabularx}
\usepackage{algorithm}
\usepackage{algpseudocode}

\newcommand{\sparagraph}[1]{\smallskip \noindent \textbf{#1}}

\title{Navigating the Shadows: Unveiling Effective Disturbances for Modern AI Content Detectors}

\author{Ying Zhou$^{1,2}$, Ben He$^{1,2}$\textsuperscript{\Letter}, Le Sun$^{2}$ \\
${}^{1}$School of Computer Science and Technology, \\
University of Chinese Academy of Sciences, Beijing, China \\
${}^{2}$Chinese Information Processing Laboratory, \\
Institute of Software, Chinese Academy of Sciences, Beijing, China \\
{\tt zhouying20@mails.ucas.ac.cn, benhe@ucas.ac.cn, sunle@iscas.ac.cn}
}

\begin{document}
\maketitle
\begin{abstract}
With the launch of ChatGPT, large language models (LLMs) have attracted global attention.
In the realm of article writing, LLMs have witnessed extensive utilization, giving rise to concerns related to intellectual property protection, personal privacy, and academic integrity.
In response, AI-text detection has emerged to distinguish between human and machine-generated content. However, recent research indicates that these detection systems often lack robustness and struggle to effectively differentiate perturbed texts. 
Currently, there is a lack of systematic evaluations regarding detection performance in real-world applications, and a comprehensive examination of perturbation techniques and detector robustness is also absent.
To bridge this gap, our work simulates real-world scenarios in both informal and professional writing, exploring the out-of-the-box performance of current detectors.
Additionally, we have constructed 12 black-box text perturbation methods to assess the robustness of current detection models across various perturbation granularities.
Furthermore, through adversarial learning experiments, we investigate the impact of perturbation data augmentation on the robustness of AI-text detectors.
We have released our code and data at \url{https://github.com/zhouying20/ai-text-detector-evaluation}.
\end{abstract}

\input{tex/introduction}
\input{tex/related_works}
\input{tex/methods}

\input{tex/attack_experiments}
\input{tex/defence_experiments}
\input{tex/conclusion}

\section*{Acknowledgements}
This work is supported by the Strategic Priority Research Program of Chinese Academy of Sciences Grant No. XDA27020200, the National Natural Science Foundation of China (62272439), and the Fundamental Research Funds for the Central Universities.
% National Natural Science Foundation of China (62272439), and the Fundamental Research Funds for the Central Universities.

\bibliography{main_refs}

\newpage
\appendix
\input{tex/appendix}

\end{document}

%% file: tex/introduction.tex
\section{Introduction}
With the rise of LLMs~\citep{DBLP:journals/corr/abs-2303-08774,DBLP:journals/corr/abs-2305-10403,DBLP:journals/corr/abs-2307-09288}, concerns about the misuse of generated content have been growing~\citep{DBLP:journals/corr/abs-2305-14552, DBLP:journals/corr/abs-2305-04812, DBLP:journals/corr/abs-2304-03738}, making AI-Text detection a topic of significant attention from the research community.
Several methods for detecting AI-generated text have recently been proposed, including fine-tuned classifiers~\citep{DBLP:conf/emnlp/UchenduLSL20,DBLP:journals/corr/abs-2306-05524}, statistical approaches~\citep{DBLP:conf/ecai/LavergneUY08,DBLP:conf/icml/Mitchell0KMF23}, watermarking~\citep{DBLP:conf/ih/AtallahRCHKMN01,DBLP:conf/icml/KirchenbauerGWK23}, and retrieval techniques~\citep{DBLP:journals/corr/abs-2303-13408}. 
Additionally, online education service providers such as Copyleak\footnote{\url{https://copyleaks.com/ai-content-detector}} and GPTZero~\citep{tian2023gptzero} have introduced AI text detection services. However, criticisms regarding misclassification results from various users have surfaced.
Simultaneously, in domains like essay writing, there is a demand from users to bypass AI text detection using perturbation methods, whereas numerous open-source tools like GPTzzz\footnote{\url{https://github.com/Declipsonator/GPTZzzs}} and AiTextDetectionBypass\footnote{\url{https://github.com/obaskly/AiTextDetectionBypass}} have emerged.
% Furthermore, inevitable minor perturbations, such as stop-word or punctuation removal, can also impact the accuracy of AI text detection, leading to certain misjudgments.
% We assert that AI-Text detection addresses a real-world application need, and it is crucial to examine the accuracy and robustness of current text detection systems in authentic writing scenarios. This evaluation is necessary as it can provide valuable insights and recommendations for subsequent detection efforts.

\input{figures/overview}

Recent efforts have begun to explore the vulnerabilities of current detection models~\citep{DBLP:journals/corr/abs-2303-14822,DBLP:journals/corr/abs-2303-11156,DBLP:journals/corr/abs-2302-05794,DBLP:journals/corr/abs-2311-08374,DBLP:conf/emnlp/ChakrabortyTZGK23}, utilizing methods such as rewrite and substitution to modify AI-generated content, rendering it indistinguishable from human-authored text. 
This underscores the importance of investigating and identifying potential weaknesses in current detectors before their deployment, ensuring their robustness and mitigating potential risks.
Simultaneously, more comprehensive work has started to summarize the issues with current detection methods and propose corresponding robustness enhancement techniques, such as RADAR~\citep{DBLP:journals/corr/abs-2307-03838} and retrieval~\citep{DBLP:journals/corr/abs-2303-13408}. 
Despite enhancing the models' defense against specific types of text perturbations to some extent, these works still face two major limitations. 
% Firstly, they primarily focus on AI text detection within specific writing scenarios, lacking evaluations of detection performance in real-world application settings. 
% Secondly, they often deal with only one type of perturbation, typically paraphrasing, without researching users' real-world needs and methods to bypass text detection.
Firstly, these efforts primarily focus on AI text detection in specific writing scenarios. Secondly, they typically involve only one type of perturbation, i.e., paraphrasing. 
In practical applications, detectors are likely to encounter a more complex and diverse set of scenarios, involving various application contexts and potential text perturbations.

% To this end, our work aims to investigate and analyze the accuracy and robustness of multiple AI text detection algorithms in real-world writing scenarios. 
To this end, our work aims to investigate and analyze the accuracy and robustness of various AI text detection algorithms in simulating real writing scenarios.
Specifically, within three categories of AI text detection methods, we evaluate six representative off-the-shelf models on data generated by ChatGPT. 
To simulate users' writing demands, we categorize AI-generated text into professional and informal writing scenarios and test detection accuracy accordingly. As expected, current text detection models exhibit lower accuracy in professional writing scenarios.
Furthermore, following an exploration of current text perturbation methods, we devise 12 types of text perturbations across four granularities. 
We apply these perturbations to the test data, generating 120,000 adversarial samples to investigate the robustness of current detection systems. 
The results reveal that, apart from the extensively studied paraphrase methods, word-level perturbations also significantly reduce AI text detection rates.
Building on earlier work, we further delve into exploring the minimum budget for adversarial learning to train robust text detectors. Additionally, we conduct preliminary investigations into transfer learning in the context of adversarial text detection.

Our work can be summarized into three parts:
1) We validate the detection accuracy of three types of current detection models in both professional and informal writing scenarios. This analysis identifies a lack of generalization performance in current detection systems.
2) We systematically and hierarchically design AI-Text perturbation methods. The results demonstrate that perturbations at various granularities significantly reduce detection performance. Additionally, we observe inconsistent performances of different detection models when faced with perturbations.
3) Budget and transfer experiments provide references and suggestions for future efforts to enhance the robustness of AI-Text detectors.

%% file: figures/overview.tex
\begin{figure}[t]
\centering
\includegraphics[scale=0.2]{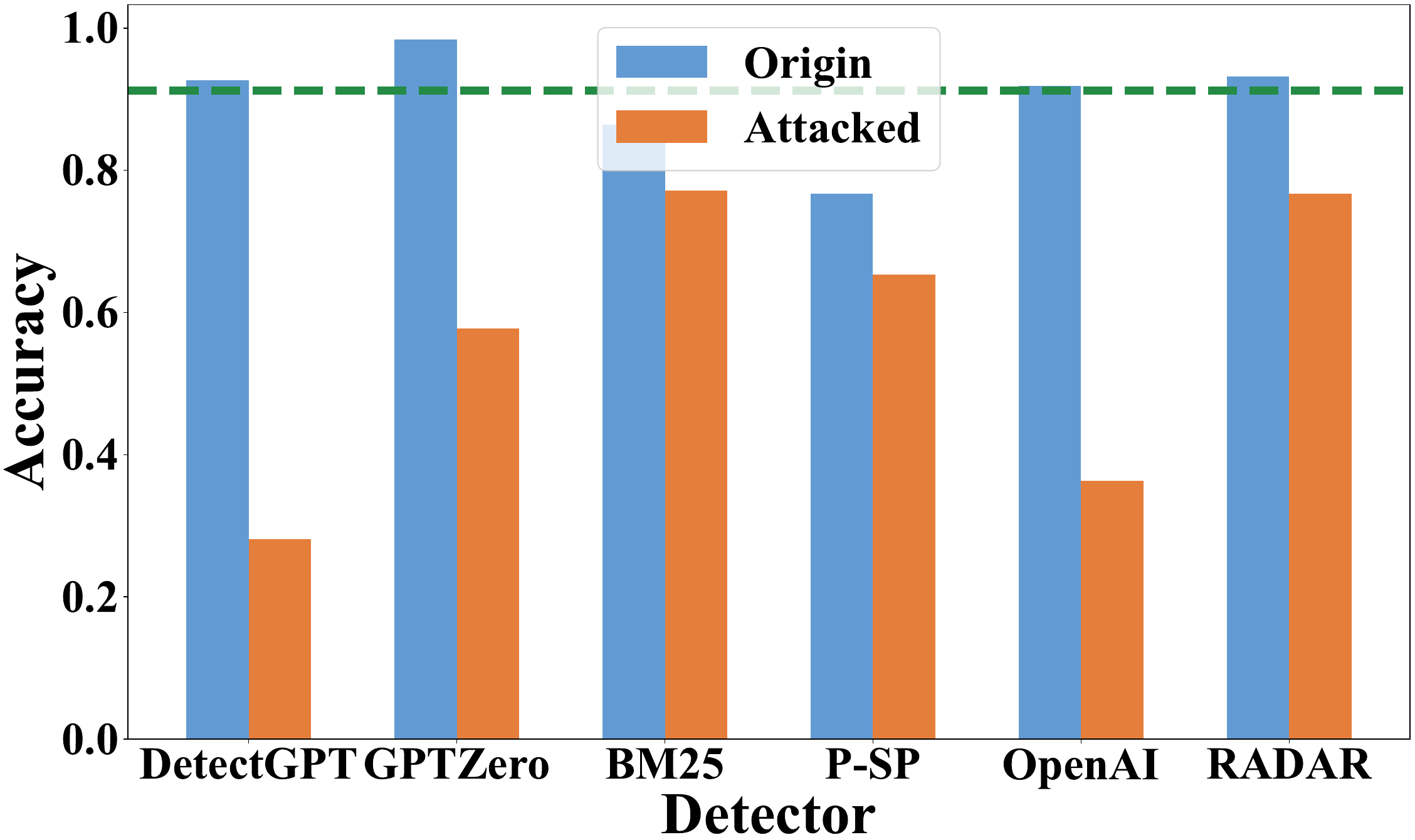}
\caption{Performance of state-of-the-art AI-text detectors significantly decreases after introducing perturbation attacks. The green dashed threshold line represents the adversarially trained RoBERTa classifier detector, achieving a detection accuracy of \textcolor[HTML]{238b45}{0.912} on the mixed test data of the original and perturbed text.} 
\label{fig:budget} 
\end{figure}

%% file: tex/related_works.tex
\section{Related Works}\label{sec:related-work}

\subsection{AI-Text Detection}
Current AI-text detectors can be categorized into four classes: 

\sparagraph{\bfseries Statistical} approaches leverage statistical tools, using metrics such as information entropy, perplexity, and \textit{n}-gram frequencies to differentiate between human and machine-generated text in a zero-shot manner~\cite{DBLP:conf/ecai/LavergneUY08,DBLP:conf/acl/GehrmannSR19,DBLP:journals/corr/abs-1908-09203,DBLP:conf/icml/Mitchell0KMF23,DBLP:journals/corr/abs-2306-05540}. Notable commercial applications include GPTZero~\citep{tian2023gptzero}, and recent open-source efforts are exemplified by DetectGPT~\cite{DBLP:conf/icml/Mitchell0KMF23}, which defines a curvature-based criterion using a log probability function for the AI detection. 

\sparagraph{\bfseries Watermark-based} methods~\citep{DBLP:conf/ih/AtallahRCHKMN01,DBLP:conf/ih/AtallahRHKSTT02,DBLP:conf/icml/KirchenbauerGWK23,DBLP:journals/corr/abs-2307-16230} is also evolving with the emergence of LLMs, where \citet{DBLP:conf/icml/KirchenbauerGWK23} randomly partition the vocabulary into a greenlist and a redlist during generation, based on the hash values of previously generated tokens.

\sparagraph{\bfseries Classifier-based} detectors~\citep{DBLP:conf/emnlp/UchenduLSL20,DBLP:journals/corr/abs-2305-16617,DBLP:journals/corr/abs-2305-09859,DBLP:journals/corr/abs-2301-07597,liu-etal-2023-coco,DBLP:journals/corr/abs-2306-05524,DBLP:conf/emnlp/WangLRJZQ23} based on supervised data typically utilize RoBERTa~\citep{DBLP:journals/corr/abs-1907-11692} to train binary classifiers for text detection. Recent efforts include OpenAI's release of detection tools~\citep{DBLP:journals/corr/abs-1908-09203}, and RADAR~\citep{DBLP:journals/corr/abs-2307-03838}, which specifically address the importance of perturbation attacks, and enhance detection robustness through adversarial learning using paraphrases.

\sparagraph{\bfseries Retrieval-based} method proposed by \citet{DBLP:journals/corr/abs-2303-13408} involves collecting historical responses from language models and assessing the AI generation likelihood of the text through semantic matching.

\subsection{Adversarial Attacks}
In addition, some studies~\citep{DBLP:journals/corr/abs-2311-08721,DBLP:journals/corr/abs-2311-08374,DBLP:journals/corr/abs-2305-10847,DBLP:journals/corr/abs-2302-05794,DBLP:journals/corr/abs-2307-02599} have addressed the impact of text perturbations on AI text detection. 
For instance, both \citet{DBLP:journals/corr/abs-2303-11156,DBLP:journals/corr/abs-2303-13408} propose to use paraphraser as the attacker to rewrite AI content, demonstrating effective attacks on many detectors. \citet{DBLP:journals/corr/abs-2306-04634} validate the detection capabilities of watermarking detectors in scenarios involving a mix of human and machine-generated text.
Furthermore, \citet{DBLP:journals/corr/abs-2305-19713} examine the significant impact of synonym perturbations on text detection performance. \citet{DBLP:conf/emnlp/KumarageSMG023} designe prompts to generate outputs more similar to human text, evading detection of existing detectors.

Notably, the recent work by \citet{DBLP:journals/corr/abs-2401-07867} has been instrumental in illustrating the susceptibility of current multilingual AI text detectors through the design of perturbations such as paraphrasing, back translation, and substitution within a multilingual context, thereby showcasing the potential benefits of adversarial training.
In contrast, our study shifts the focus towards the detectability of AI-generated text in practical scenarios. We utilize AI-generated text outputs that more closely mimic human-produced content, develop a broader range of perturbation attacks, and critically, expand our examination beyond the conventional classifier-based methods. Our evaluation includes not only classifiers but also involves retrieval systems and other detection mechanisms, thereby providing a more holistic assessment of detection efficacy in diverse operational environments.
%In comparison to \citet{DBLP:journals/corr/abs-2401-07867}, our study concentrates on the detectability of AI-generated text in real-world scenarios. We employ AI-generated text outputs that closely resemble human output, design a more comprehensive set of perturbation attacks, and importantly, extend our focus beyond simple classifier methods. We evaluate the detection performance not only for classifiers but also for retrieval and other detection tools.

%% file: tex/methods.tex
\section{Experimental Setup}
In this section, we first survey the current state-of-the-art AI-text detection methods.
Subsequently, considering the presence of intentional or unintentional perturbation attacks in real-world applications that can impact the performance of detection models, we synthesize and implement 12 black-box perturbation methods. Here, ``black-box'' refers to attacking algorithms lacking access to internal information of detectors, such as gradients or hidden states. 
Meanwhile, building upon the scoring-based configuration of existing detectors, we investigate the challenges associated with metric selection and threshold determination in evaluation.

\subsection{Off-the-Shelf Detectors}
As described in Section \ref{sec:related-work}, the current research in AI detection primarily focuses on four directions. 
However, the application of watermarking techniques to commercial or open-source LLMs remains limited, with few practical implementations to date. Consequently, our investigation focuses on three types of readily deployable detection models:

\begin{enumerate}[leftmargin=*, itemsep=-0.2em, topsep=0.2em]
\item Statistical models, i.e., DetectGPT~\citep{DBLP:conf/icml/Mitchell0KMF23} and GPTZero~\citep{tian2023gptzero};
\item Retrieval-based models~\citep{DBLP:journals/corr/abs-2303-13408} including BM25~\citep{DBLP:conf/trec/RobertsonWHGP95} and P-SP~\citep{DBLP:conf/emnlp/WietingGNB22};
\item Classifier models like OpenAI's text classifier~\citep{DBLP:journals/corr/abs-1908-09203} and RADAR~\citep{DBLP:journals/corr/abs-2307-03838}.
\end{enumerate}

\noindent Additionally, to accurately assess the impact of training data on classifier detectors, we follow OpenAI's approach to train a RoBERTa-base as a comparative baseline on the two datasets we employed. Furthermore, considering the dependence of retrieval models on corpus data, we also evaluate the influence of documents from four different sources on detection performance. The specific details will be elaborated in Section~\ref{sec:exp-detectability}. In summary, we assessed a total of 6 off-the-shelf detection models and expanded our evaluation to cover 13 experimental settings.

\subsection{Adversarial Attacks}
To simulate real-world scenarios where users may modify AI-generated text for cheating purposes and also to account for noise in information transmission, we devised 12 perturbation attack methods across four granularities, i.e., document, sentence, word, and character. 
Several of our attack strategies build on the foundations laid by previous research, as evidenced by studies in~\citep{DBLP:journals/corr/abs-2310-14724,DBLP:journals/corr/abs-2307-02599,DBLP:journals/corr/abs-2303-13408,DBLP:journals/corr/abs-2305-19713,DBLP:journals/corr/abs-2303-14822}, while others are first introduced in this work, representing a novel exploration of their effect on the detectability of AI-generated text.

\subsubsection{Document-level Perturbations}
\sparagraph{Paraphrase.} We employ the highly effective DIPPER~\citep{DBLP:journals/corr/abs-2303-13408} rewriter with the lex=40, order=40, which is the most intensive settings in their paper.

\sparagraph{Back-Translation.} Leveraging Neural Machine Translation (NMT) models, we choose French as the intermediary language, and utilized the translation models from Helsinki-NLP~\citep{DBLP:conf/eamt/TiedemannT20}.

\subsubsection{Sentence-level Perturbations}
\sparagraph{Sentence Back-Translation.} Akin to document-level Back Translation, but randomly selecting sentence windows for translation. Up to 3 pieces are perturbed within a maximum window of 5 sentences.

\sparagraph{MLM Prediction.} Randomly masking 2 to 5 sentences in the original text and replacing them using the BART-large~\citep{DBLP:conf/acl/LewisLGGMLSZ20} model. 
% Each document undergoes random perturbation of  sentences.

\subsubsection{Word-level Perturbations}
\sparagraph{MLM Prediction for Words.} Akin to the sentence MLM prediction, using the BERT-base~\citep{DBLP:conf/naacl/DevlinCLT19} model to replace random tokens with synonyms. To control text quality, the maximum word perturbation ratio per article does not exceed 20\%. This setting is also applied to all our word-level perturbations.

\sparagraph{Adverb Insertion.} Randomly inserting a relevant adverb before verbs in the original text.

\sparagraph{Spelling Errors.} Simulating situations where users misspell words due to ignorance, implemented through a predefined spelling error dictionary.

\sparagraph{Keyboard Typos.} Simulating typos during keyboard input, including substitution of nearby characters, swapping adjacent characters, inserting irrelevant characters, and deleting specific characters.

\subsubsection{Character-level perturbations.}
\sparagraph{Word Merging.} Simulating scenarios in information transmission contexts where spaces between words are missing. Introducing 3-10 randomly chosen word merging errors per article.

\sparagraph{Case of the First Character of a Word.} Simulating scenarios where the first character of a word is incorrectly capitalized.

\sparagraph{Punctuation Removal.} Simulating that punctuation is lost, randomly removing up to 30\% of punctuation marks from the original text.

\sparagraph{Space Insertion.} Building upon prior work~\citep{DBLP:journals/corr/abs-2307-02599}, we control the insertion of spaces to between 5-10 spaces per article.

\input{tables/dataset}

\subsection{Evaluation Metrics}
\sparagraph{Detection.} The prevailing practice in current research is to use the AUC-ROC to comprehensively evaluate the discriminative capability of detectors for AI-generated text~\citep{DBLP:conf/icml/Mitchell0KMF23,DBLP:conf/icml/KirchenbauerGWK23}.
However, in the real-world deployment of AI-text detector, it is essential to select a fixed threshold based on training strategies and test data to support subsequent detection, e.g., GPTZero considers probabilities greater than 0.88 as ``Entirely AI.''. 
The threshold-independent AUC-ROC metric may no longer accurately reflect the detection performance in practical tests.
Therefore, we opt for {\bfseries F1} and {\bfseries Accuracy} metrics to assess how accurately input texts are detected as AI-generated content.
As F1 scores are heavily influenced by the chosen detection threshold, we calibrate the threshold by maximizing Youden's J statistic for each detection method on a reserved set of 5000 samples.
This threshold is then fixed to validate model robustness under perturbations.

\sparagraph{Robustness.} In perturbation attack experiments, we consider the {\bfseries Attack Success Rate (ASR)} as the metric, i.e., the accuracy change for AI text detection after perturbation.
% and 2) the change in F1 scores before and after the attack, denoted as $\Delta F1$, serving as another metric to assess the extent of model detection capability degradation.

% \paragraph{Misc.}

\input{tables/detection}

\subsection{Benchmarks}
As mentioned earlier, this paper aims to validate the detectability of AI-generated text in real-world scenarios, focusing specifically on the most successful commercial LLMs, the GPT series~\citep{radford2019language,DBLP:conf/nips/BrownMRSKDNSSAA20,DBLP:conf/nips/Ouyang0JAWMZASR22}. In contrast to previous work, our attention is solely on data generated by the ChatGPT\footnote{\url{https://chat.openai.com}}, which is readily accessible to the end users.
We employ two datasets in the experiments. {\bfseries CheckGPT}~\citep{DBLP:journals/corr/abs-2306-05524} centers around professional writing, which consists of a dataset of 900 thousand samples encompassing news articles, essays, and scientific research generated using various prompts. {\bfseries HC3}~\citep{DBLP:journals/corr/abs-2301-07597} focuses on internet QA scenarios, employing the continuation-writing method to generate ChatGPT responses in fields such as encyclopedia, community, finance, medicine, and open-ended questions. 
Through these two datasets, we simulate the text detection needs of both professional and ordinary users, with detailed information on the two datasets provided in Table~\ref{tab:data}.
As for adversarial attack experiments, we generate large-scale perturbed datasets based on the attack methods described above, resulting in 1.08 million perturbed samples for CheckGPT, and 192 thousand perturbations for HC3.

\subsection{Research Questions}
Based on off-the-shelf detectors, publicly available data, and black-box perturbations, we propose three research questions to investigate whether current AI-text detectors' development can meet the demands of various real-world application scenarios:

\begin{itemize}[leftmargin=*, itemsep=-0.2em, topsep=0.2em]
\item {\bfseries RQ1.} What is the detection accuracy when applying current detectors directly to the SoTA LLM-generated texts?
\item {\bfseries RQ2.} How does the performance of current detection systems change when facing different perturbations? What are the most effective attack methods?
\item {\bfseries RQ3.} When facing perturbation attacks, can the training strategy or settings of the detection system be adjusted to achieve robust detection?
\end{itemize}

\noindent In the following sections, we will address RQ1 and RQ2 in Section~\ref{sec:detect} by evaluating the detectors in real-world scenarios. In Section~\ref{sec:robust}, we will explore adversarial learning methods to enhance the robustness of current classifier-based detectors.
%to provide feasible research directions for future work.

%% file: tables/dataset.tex
\begin{table}[t]
\centering         % had to be here!
\resizebox{0.99\linewidth}{!} {%
\begin{tabular}{l|*{2}{c}}
\toprule
& {\bfseries CheckGPT} & {\bfseries HC3}\\
\midrule
Train data & 720,000* & 58,508 \\
Test data & 90,000* & 25,049 \\
Avg \#words  & 136.68 & 145.89 \\
Domain & News, Essay, Research & QA \\
\bottomrule
\end{tabular}
}
\caption{Data statistics, where * denotes the data are randomly split with seed 42, and \#words denotes the number of words in one sample.}
\label{tab:data}
\end{table}

%% file: tables/detection.tex
\begin{table*}[t]
\centering         % had to be here!
% \resizebox{0.99\linewidth}{!} {%
\begin{tabular}{l *{4}{c} | *{4}{c}}
\toprule
\multirow{2}{*}{\bfseries Detectors} & \multicolumn{4}{c}{\bfseries Professional Writing} & \multicolumn{4}{c}{\bfseries Informal Writing} \\
\cmidrule{2-5} \cmidrule{6-9}
& {\bfseries F1} & {\bfseries Acc$_{G}$} & {\bfseries Acc$_{H}$} & {\bfseries Thres.}
& {\bfseries F1} & {\bfseries Acc$_{G}$} & {\bfseries Acc$_{H}$} & {\bfseries Thres.} \\
\midrule
{DetectGPT} & 73.30 & 71.23 & 76.81 & 0.271 & 90.95 & 92.64 & 89.16 & 0.579 \\
{GPTZero} & 90.12 & 86.90 & 93.95 & 0.572 & 99.17 & 98.35 & 100.0 & 0.443 \\
\midrule
{BM25$_{Train}$} & 55.39 & 45.94 & 80.02 & 0.321 & 85.65 & 86.41 & 84.97 & 0.288 \\
{BM25$_{Train^+}$} & 97.78 & 98.32 & 97.20 & 0.604 & 98.49 & 98.91 & 98.10 & 0.392 \\
{BM25$_{ShareGPT}$} & 40.44 & 29.64 & 82.98 & 0.243 & 78.60 & 77.95 & 80.06 & 0.221 \\
{BM25$_{ShareGPT^+}$} & 98.21 & 98.36 & 98.04 & 0.434 & 98.49 & 98.83 & 98.18 & 0.373 \\
\midrule
{OpenAI} & 64.46 & 55.33 & 83.62 & 0.071 & 93.90 & 91.91 & 96.24 & 0.829 \\
{RADAR} & 72.23 & 69.28 & 77.41 & 0.306 & 69.36 & 93.20 & 26.11 & 0.354 \\
{RoBERTa} & 98.96 & 98.56 & 99.36 & 0.943 & 99.80 & 99.96 & 99.64 & 0.942 \\
\bottomrule
\end{tabular}
% }
\caption{Detection performance of off-the-shelf detectors on CheckGPT and HC3 datasets.
Acc$_G$: detect accuracy of GPT-generated text.
Acc$_H$: detect accuracy of human-written text.
Thres: the threshold determined by maximizing Youden's J statistic.
%, hence, this detection performance can be considered as the optimal performance of the detector on the current test data.
}
\label{tab:detection}
\end{table*}

%% file: tex/attack_experiments.tex
\section{Evaluating Detectors in the Wild}\label{sec:detect}

\input{tables/detect-gpt2}
\input{tables/attack}
\input{tables/quality}

\subsection{Detectability of the Cutting-Edge AI-Text}\label{sec:exp-detectability}
We initially validate the performance of three types of AI text detection algorithms on cutting-edge AI text datasets. In our experiments, we consider the HC3 dataset, derived from internet-based QA data, as representative of informal writing scenarios, and the CheckGPT dataset, based on academic paper writing, as representative of professional writing scenarios.

\sparagraph{AI-texts are more easily detected in informal writing scenarios.} As shown in Table~\ref{tab:detection}, almost all detectors exhibit higher false positives in professional writing contexts compared to informal writing contexts. 
Taking the commercial detection tool GPTZero as an example, it demonstrates minimal false positives in informal writing scenarios, showcasing strong practical utility. However, in CheckGPT, the performance has significantly declined, where the F1 score dropped from 99.2 to 90.1, markedly lower than the finetuned RoBERTa's 98.9. 
Surprisingly, the adversarially trained RADAR model exhibits severe false positives in informal writing scenarios, possibly stemming from partial overlap in training data between RADAR and HC3 datasets. This overlap may lead to overfitting to the paraphraser on which the model relies, making it challenging to distinguish human-generated text in that particular domain.

\sparagraph{The retrieval method heavily relies on the test samples within the document corpus.} As for the retrieval method proposed by \citet{DBLP:journals/corr/abs-2303-13408}, we conduct ablation experiments on its corpus data. As seen in Table~\ref{tab:detection}, taking the CheckGPT dataset as an example, when utilizing only the training data of the RoBERTa detector or publicly available ShareGPT
data, namely BM25$_{Train}$ and BM25$_{ShareGPT}$, the retrieval method exhibits the poorest performance, struggling to distinguish AI-text. 
However, upon incorporating the test data into the retrieval corpus, i.e., BM25$_{Train^+}$ and BM25$_{ShareGPT^+}$, the accuracy rapidly improves to over 98\%, as every machine-generated text now shares identical retrieval results. 
This performance poses a significant challenge in practical applications, as providers of retrieval detection services must be capable of acquiring and storing all generated results of target LLMs.
Efficiency, security, privacy, and other related concerns may limit the widespread adoption of such retrieval detection.

\sparagraph{Classifiers-based detectors exhibit poor generalization performance.} OpenAI, RADAR, and the fine-tuned RoBERTa model can be considered as three models with the same architecture, with training data quality continually improving. Specifically, each model is trained on data generated by GPT-2, Vicuna, and ChatGPT, respectively. 
Excluding RADAR's human accuracy on HC3 data, based on GPT detection performance, it is evident that the quality of training data for classifier-based detectors positively correlates with AI text detection performance on cutting-edge AI-generated content.
Furthermore, as shown in Table~\ref{tab:detection-gpt2}, the OpenAI detector performs poorly on ChatGPT data, and the RoBERTa trained on ChatGPT data exhibits suboptimal detection performance on GPT-2 text.
These results indicate that neural network-based AI text detectors have limited generalization performance. When the testing data differs in generation methods, model scale, and other aspects from the training data, the model's detection performance sharply declines.

\subsection{Effectiveness of Perturbations}
We further delve into perturbation scenarios, examining the impact of intentional or unintentional text perturbations generated by users using AI tools on the performance of detectors. Specifically, we investigate the extent of the decline in detection accuracy for AI-generated text across four levels of perturbation granularity.

\sparagraph{All detectors exhibit vulnerability to perturbations, even after defense training.} From Table~\ref{tab:attack}, it is evident that all detectors show significant misjudgments in the presence of text perturbations, with an average ASR exceeding 10\%. 
Among them, the retrieval and the RADAR methods, proposed for robustness issues, demonstrate a certain degree of defensive performance. However, when facing specific perturbation attacks, they still exhibit weaker detection capabilities. 
For instance, the retrieval method, due to its ability to access the original AI-generated text on the test set, shows high defense capabilities against minor text perturbations such as typos and spaces. 
However, its defense capability sharply declines in scenarios involving substantial deviations from the original text, such as rewriting and back translation. Furthermore, as seen in Table~\ref{tab:attack_retrieval}, once the retrieval method cannot access the test set, its detection performance and robustness significantly decrease.
As for RADAR, based on paraphrasing for adversarial training, it exhibits a strong defense against larger granularity perturbations. Nevertheless, it inherits the vulnerability of neural network models and performs poorly on perturbations at the word level.
A similar performance could also be observed on the HC3 dataset in Table~\ref{tab:attack_hc3}.

\sparagraph{Statistical and classifier-based methods exhibit similar performance when facing perturbations.} From Table~\ref{tab:attack}, we observe that, whether it is the commercial GPTZero or other open-source detectors, introducing word-level perturbations to AI-generated articles yields more significant attack results compared to full-text rewriting.
Moreover, the effectiveness of word-level perturbation methods appears consistent across both groups. For instance, both MLM word substitution and spelling errors lead to higher attack success rates in all statistical and classifier-based models. 
This may imply a greater reliance on statistical metrics, such as perplexity, in the current classifier training. Future work could focus on improving these aspects.

\sparagraph{Perturbed texts show significant changes in text quality, readability, or semantic similarity.} To assess the changes in semantic similarity and readability introduced by perturbed text, we report four text quality metrics. 
1) The semantic similarity between the original and perturbed text, calculated using the P-SP model~\citep{DBLP:conf/emnlp/WietingGNB22}. 
2) The Flesch Reading Ease score, quantifying text readability, with 0 indicating a highly specialized text and 100 representing a fifth-grade level. 
3) Text quality scores judged by GPT-3.5-Turbo, ranging from 0 to 10, with 10 being the highest score. The specific prompt used is provided in Appendix~\ref{sec:appendix-judge}. 
4) Perplexity, assessed using the 7B LLaMA-2-base model~\citep{DBLP:journals/corr/abs-2307-09288} to evaluate text fluency. 
From Table~\ref{tab:quality}, it is evident that the success rate of text perturbation is inversely correlated with text quality to a certain extent. Perturbation methods such as Typos can even decrease the GPT score from 8.85 to 3.95.

\subsection{Discussion on RQ1\&2}
In summary, for RQ1 and RQ2, we can learn from the results that the detection methods based on statistical metrics are generally applicable in informal scenarios. Their zero-shot characteristics endow them with a certain degree of generalization ability. 
When targeting a certain LLM, training a classifier-based detector, given sufficient training data, proves to be a viable option. However, its generalization capability to other LLMs may be limited. 
In scenarios with substantial perturbations, retrieval methods exhibit the strongest defense capabilities. Nevertheless, their reliance on the original generated text may constrain their applicability. 
When data from the same distribution is unavailable, both their detection and defense performance significantly decline. For details, please refer to Table~\ref{tab:attack_retrieval} in the Appendix.
In future research, proposing more robust detection models or strategies that blend current detection system outcomes would be worthwhile directions.

%% file: tables/detect-gpt2.tex
\begin{table}[t]
\centering         % had to be here!
% \resizebox{0.99\linewidth}{!} {%
\begin{tabular}{l *{2}{c} *{3}{c}}
\toprule
{\bfseries Datasets} & {\bfseries OpenAI} & {\bfseries RoBERTa} \\
\midrule
{GPT-2-Small} & 97.29 & {\bfseries 57.85} \\
{GPT-2-Medium} & 96.96 & {\bfseries 63.07} \\
{GPT-2-Large} & 96.74 & {\bfseries 65.59} \\
{GPT-2-XL} & 95.35 & {\bfseries 65.62} \\
{HC3} & 93.90 & 99.80 \\
{CheckGPT} & {\bfseries 64.46} & 98.96 \\
\bottomrule
\end{tabular}
% }
\caption{F1 scores for OpenAI detector trained on GPT-2 data and our RoBERTa detector trained on ChatGPT data on both test sets. Lower F1 scores are in {\bfseries bold}.}
\label{tab:detection-gpt2}
\end{table}

%% file: tables/attack.tex
\begin{table*}[t]
\centering         % had to be here!
% \resizebox{\linewidth}{!}{%
\begin{tabular}{ @{} ll *{6}{c} @{} }
\toprule
\multicolumn{2}{c}{\multirow{2}{*}{{\bfseries Perturbations}}} & \multicolumn{2}{c}{\bfseries Statistic} & {\bfseries Retrieval} & \multicolumn{3}{c}{\bfseries Classifier} \\
\cmidrule(lr){3-4} \cmidrule(lr){5-5} \cmidrule(lr){6-8}
\multicolumn{2}{c}{} & {\bfseries DetectGPT} & {\bfseries GPTZero} & {\bfseries BM25$_{Train^+}$} & {\bfseries OpenAI} & {\bfseries RADAR} & {\bfseries RoBERTa} \\
\midrule
& {\bfseries Origin F1} & 73.30 & 90.12 & 97.78 & 64.46 & 72.23 & 98.96 \\
\midrule
\multirow{2}{*}{Doc} & Paraphrase & {\bfseries 29.09} & {\bfseries 41.67} & {\bfseries 67.16} & 4.79 & 3.24 & {\bfseries 66.24} \\
& BackTrans & {\bfseries 38.11} & 19.05 & {\bfseries 43.67} & 8.23 & 0.76 & {\bfseries 25.93} \\
\midrule
\multirow{2}{*}{Sent} & BackTrans & {\bfseries 30.04} & 14.29 & 12.98 & 8.23 & 1.48 & 12.62 \\
& MLM & 14.70 & {\bfseries 39.29} & {\bfseries 22.29} & 2.36 & 2.48 & 12.66 \\
\midrule
\multirow{4}{*}{Word}
& MLM & {\bfseries 68.88} & {\bfseries 83.73} & 4.39 & 19.30 & 2.12 & {\bfseries 75.59} \\
& AdvInsert & {\bfseries 64.20} & {\bfseries 71.43} & 0.00 & {\bfseries 31.56} & {\bfseries 25.93} & {\bfseries 47.26} \\
& Spelling & {\bfseries 70.48} & {\bfseries 62.70} & 0.00 & {\bfseries 52.62} & {\bfseries 29.92} & {\bfseries 87.10} \\
& Typos & {\bfseries 70.95} & {\bfseries 36.51} & 0.00 & {\bfseries 54.25} & {\bfseries 38.31} & {\bfseries 64.68} \\
\midrule
\multirow{4}{*}{Char} 
& Merge & 17.82 & {\bfseries 23.81} & 0.00 & {\bfseries 45.83} & 2.60 & {\bfseries 27.85} \\
& Case & {\bfseries 44.39} & {\bfseries 80.16} & 0.00 & {\bfseries 52.22} & 14.38 & {\bfseries 39.63} \\
& Punctuation & {\bfseries 23.13} & {\bfseries 25.00} & 0.00 & {\bfseries 29.76} & 0.28 & 10.11 \\
& SpaceInsert & {\bfseries 35.36} & 11.51 & 0.00 & {\bfseries 52.86} & 1.60 & {\bfseries 21.45} \\
\midrule
& {\bfseries Average ASR} & 42.26 & 42.43 & 12.54 & 30.17 & 10.26 & 40.93 \\
\bottomrule
\end{tabular}
% }
\caption{Attack Success Rates (ASR) of perturbations on the CheckGPT test set. 
% The Retrieval method utilizes training and test data as retrieval documents, and the threshold for all detection algorithms is set to the optimal result on the original test data to simulate real-world model deployment scenarios. 
A higher ASR indicates a higher proportion of AI-generated text misclassified as human text after perturbation. 
All ASR exceeding {\bfseries 20\%} are highlighted in {\bfseries bold}.
}
\label{tab:attack}
\end{table*}

%% file: tables/quality.tex
\begin{table}[t]
\centering         % had to be here!
\resizebox{\linewidth}{!}{%
\begin{tabular}{ @{} l *{4}{c} @{} }
\toprule
& {\bfseries Sim~$\uparrow$} & {\bfseries Flesch} & {\bfseries GPT~$\uparrow$} & {\bfseries PPL~$\downarrow$} \\
\midrule
Origin & 100.0 & 26.55 & 8.85 & 6.18 \\
\midrule
Paraphrase & 80.51 & 35.91 & 7.38 & 9.75 \\
BackTrans & 86.23 & 16.62 & 6.93 & 20.18 \\
\midrule
BackTrans & 92.13 & 25.87 & 7.91 & 9.98 \\
MLM & 81.90 & 36.23 & 4.73 & 8.71 \\
\midrule
MLM & 67.16 & 37.34 & 3.00 & 29.81 \\
AdvInsert & 97.98 & 20.38 & 4.29 & 12.71 \\
Spelling & 87.32 & 29.08 & 3.49 & 24.55 \\
Typos & 80.38 & 29.97 & 3.95 & 23.14 \\
\midrule
Merge & 98.77 & 20.43 & 8.81 & 8.04 \\
Case & 99.81 & 26.61 & 7.10 & 10.06 \\
Punctuation & 99.49 & 19.31 & 8.24 & 7.49 \\
SpaceInsert & 97.03 & 30.55 & 8.18 & 8.99 \\
\bottomrule
\end{tabular}
}
\caption{Comparative results of the quality between original and perturbed text. An upper arrow indicates that higher values are desirable, and vice versa. A higher Flesch value signifies more easily understandable text.}
\label{tab:quality}
\end{table}

%% file: tex/defence_experiments.tex
\section{Robustness Enhancement}\label{sec:robust}

\subsection{Defence Budgets}
To further investigate the role of perturbed sample augmentation in enhancing the robustness of AI text detectors, we conducte experiments to evaluate the performance variation of the adversarially trained RoBERTa detector under different perturbation budgets.
We define the perturbation budget in two aspects: firstly, the number of augmented samples for each perturbation during adversarial training; and secondly, the transferability of different perturbation methods under the same granularity.
In this study, we choose the RoBERTa model trained on the CheckGPT dataset as our test setting.
The results of these two aspects are illustrated in Figure~\ref{fig:budget} and Table~\ref{tab:transferability}.

\input{figures/budget}
\input{tables/transferability}

\sparagraph{3,000 Perturbed Samples is All You Need.} From Figure~\ref{fig:budget}, we observe the impact of the number of perturbed samples used as augmentation data during the fine-tuning of the RoBERTa model on the average ASR.
Our results demonstrate that incorporating a small number of perturbed samples effectively enhances the model's defensive capability against these perturbations.
This increasing trend plateaus when the number of perturbed samples reaches around 3000, showing a gradual decline.
Ultimately, with the addition of 10,000 perturbed samples (12 perturbation methods, totaling 120,000 augmented data), the average attack success rate decreases from 40.93 to 8.01.

\sparagraph{Defense capabilities obtained through transfer learning are not stable.} As for transferability, we selected Paraphrase, MLM-Sentence, MLM-Word, and Space Inserting as target perturbations for each of the four granularities. 
For each experiment, one perturbation is reserved as the target, while the remaining 11 perturbations are used for adversarial training.
We evaluate the detector's defensive capability against the target perturbation post-adversarial training, and the experimental results are presented in Table~\ref{tab:transferability}.
After fine-tuning, there was a significant decrease in in-domain ASR across the 11 perturbation data, all falling below 9\%.
However, for out-of-distribution (OOD) target perturbations, notable differences can be observed.
The MLM-Sentence method, which is more amenable to transfer learning, exhibits a substantial 65.8 decrease in ASR without specific training, with an ASR of only 9.79. 
In contrast, the more challenging MLM-Word achieves only 3.8 in transfer performance and maintains a high ASR of 43.47 post-training.
These results suggest that relying on transfer learning alone to address the robustness of AI text detection is not realistic. Subsequent work should consider a more comprehensive coverage of perturbation attacks.

\subsection{Discussions on RQ3}
To summarize RQ3, concerning text perturbations, augmenting the training data with perturbed samples can enhance the robustness of the detector to some extent.
However, there is an upper limit to this enhancement, and the trend levels off after 3,000 perturbed samples. 
Meanwhile, vanilla transfer learning for defense brings about unstable improvements, depending on whether the target perturbation patterns can be learned from the other in-domain perturbation methods.

%% file: figures/budget.tex
\begin{figure}[t]
\centering
\includegraphics[scale=0.135]{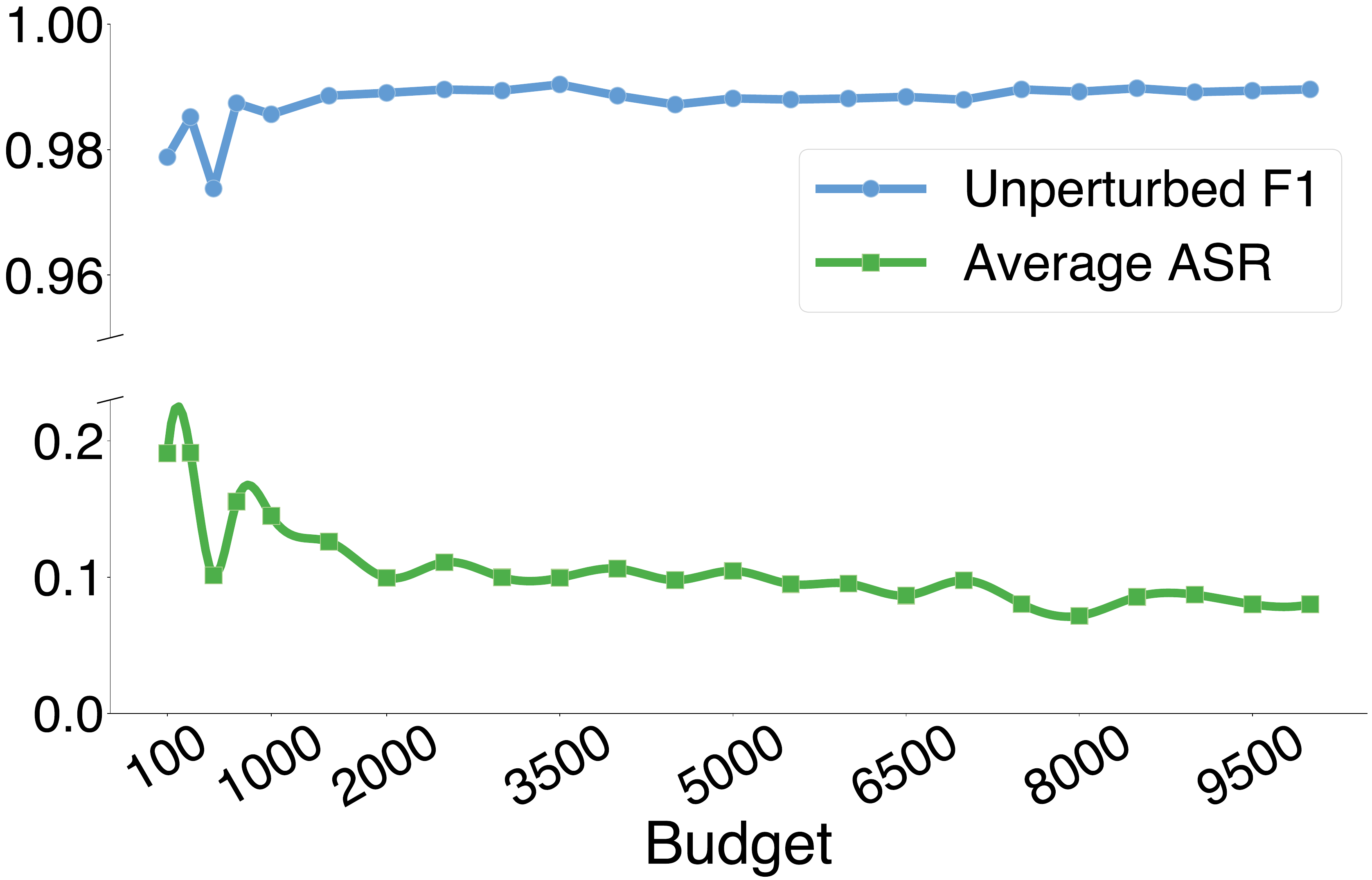}
\caption{Gradual reduction in average ASR with an increase in the number of perturbed data augmentations. Meanwhile, the F1 score on unperturbed data remains relatively stable, around 0.98. Refer to Appendix~\ref{sec:appendix-exp} for details.} 
\label{fig:budget} 
\end{figure}

%% file: tables/transferability.tex
\begin{table}[t]
\centering         % had to be here!
% \resizebox{0.99\linewidth}{!} {%
\begin{tabular}{l *{2}{c}}
\toprule
& {\bfseries In-domain ASR} & {\bfseries OOD $\Delta$ASR} \\
\midrule
Paraphrase & 4.82 & -29.92 \\
MLM-Sent & 8.52 & {\bfseries -65.80} \\
MLM-Word & 7.98 & \underline{-3.80} \\
Space-Insert & 7.90 & -11.71 \\
\bottomrule
\end{tabular}
% }
\caption{Transfer learning results for perturbation attacks. $\Delta$ASR represents the reduction in ASR on that target perturbation after training.}
\label{tab:transferability}
\end{table}

%% file: tex/conclusion.tex
\section{Conclusions}
In this paper, we study two real-world application scenarios for AI text detection: professional writing and informal writing. 
We evaluate the current SoTA detection performance in both scenarios using three categories of detection methods and six representative models. %The results indicate that zero-shot detectors struggle with more complex professional writing scenarios. Additionally, we observe a lack of generalization ability of classifier-based detectors to non-training data.
Furthermore, we introduce and design a set of 12 text perturbation methods, demonstrating the vulnerability of current detection models at different levels of granularity. %We conduct a thorough analysis of the perturbed text quality.
Finally, we apply adversarial learning in the context of perturbed data augmentation, validating the minimum budget and transferability of enhancing classifier models. %Our experimental results reveal limitations and challenges in current AI text detection.
In future work, we plan to extend our evaluations to include more LLM-generated data, such as Vicuna~\citep{vicuna2023} and Mistral~\citep{DBLP:journals/corr/abs-2310-06825}.

\section*{Limitations}
This paper aspires to provide a comprehensive evaluation and analysis of the overall performance of state-of-the-art AI detectors. 
However, given the challenges posed by multilingual and multi-modal applications, our study may not fully cover all aspects. Additionally, it is acknowledged that we cannot encompass all existing text perturbation methods, and the 4 levels of granularity and 12 perturbation tools we construct may not entirely cover real-world scenarios. 
Thus, the definition and evaluation of real-world application scenarios in this paper may lack more comprehensive coverage and consideration.
Furthermore, this work focuses on adversarial learning to improve the robustness of classifier-based detectors and does not delve into designing more complex and effective defense algorithms.
Considering the rapid development of bypass methods for AI-text detectors in reality, more in-depth research on the robustness of AI detection may be a direction for future work.

\section*{Ethics Statement}
In this paper, we explore the detectability of AI-text in professional and informal writing scenarios and validate the vulnerabilities in current detection systems through perturbation experiments.
Our aim is to provide insights and recommendations for the design and training of robust AI detection frameworks in subsequent research. 
Additionally, we offer robustness validation methods to facilitate the reliable deployment of detection systems for commercial use.

%% file: tex/appendix.tex
\section{Detailed Results of Perturbations} \label{sec:appendix-exp}

\input{tables/appendix-word-proportion}

\input{tables/appendix-origin-detect-budget}

\paragraph{Proportion of word-substitution.} For word substitution perturbations, we analyzed different levels of perturbation by varying the proportion of perturbed tokens within the entire text. As shown in Table~\ref{tab:word_sub_proportion}, we evaluated the ASR on the RoBERTa classifier using the CheckGPT dataset. We can see that, under 1\% perturbation (where each article is perturbed by only one to two words), the attack success rate is approximately 1\%. As the perturbation reaches 7\% (averaging around 10 perturbed words per article), the ASR for all three word substitution methods exceeds 10\%.

\paragraph{Detection on unperturbed data.} Additionally, we provide supplementary data for Figure~\ref{fig:budget} in Table~\ref{tab:adv_f1_human}. The adversarially trained model shows improved defense against perturbed data without compromising detection performance on unperturbed text under various data augmentation budgets.

\paragraph{Cost of Attacking}
We spend about 5000 GPU hours on A100 GPUs for generating the perturbed datasets and evaluating the off-the-shelf detectors.

\input{tables/appendix-attack-hc3}

\input{tables/appendix-attack-retrieval}

\section{GPT Judgement Prompt} \label{sec:appendix-judge}
Following the GPT judgement method proposed by \citet{DBLP:journals/corr/abs-2307-03838}, we conducted scoring experiments on 2,503 AI-generated texts from the CheckGPT dataset using the GPT-3.5-Turbo API. The prompts for both original and perturbed texts were as follows:
\textit{You are given an array of 13 sentences. Please rate these sentences and reply with an array of scores assigned to these sentences. Each score is on a scale from 1 to 10, the higher the score, the sentence is written more like a human. Your reply example: [2,2,2,2,2,2,2,2,2,2,2,2,2].}

\section{Perturbation Samples}
In this section, we show the original AI-generated sample and all perturbed texts for a random sample.

\paragraph{Origin.} \textit{In this paper, we explore grand unified theories that utilize an SU(5)xSU(5) gauge group. Our focus is on preventing fast proton decay through a combination of small triplet couplings and a large triplet mass, achieved through discrete symmetries. We demonstrate that in many of our models, the GUT scale ($M_{GUT}$) occurs naturally due to a balance of higher dimension terms and soft supersymmetry breaking masses. Our findings include intriguing patterns in quark and lepton masses, and we examine the differences between grand unified theories and string unification.}

\paragraph{Paraphrase.} \textit{Here we look at Grand Unified Theories which make use of the SU(5)xSU(5) gauge group, concentrating on avoiding fast proton decay by the use of small triplet couplings and large triplet masses, obtained through discrete symmetries. We show that in many of our models, the GUT scale ($M_{GUT}$) arises naturally from a balance between higher dimension terms and the soft breaking of supersymmetry. We find some unusual patterns in the quark and leptoon masses, and we also discuss the differences between the GUT approach and the string approach to unified theories.}

\paragraph{Back Translation.} \textit{In this paper, we examine the main unified theories that use a SU(5)xSU(5) gauge group. We focus on preventing the rapid decomposition of protons by a combination of small triplet couplings and large triplet mass obtained by discrete symmetries. We show that in many of our models, the GUT ($M_{GUT}$) scale occurs naturally due to a balance of upper dimensional terms and soft supersymmetry break masses.}

\paragraph{Back Translation Sentence.} \textit{In this paper, we examine the main unified theories that use a SU(5)xSU(5) gauge group. We focus on preventing the rapid decomposition of protons by a combination of small triplet couplings and large triplet mass obtained by discrete symmetries. We show that in many of our models, the GUT scale ($M_{GUT}$) occurs naturally due to a balance of the upper dimension terms and the soft supersymmetry break masses.}

\sparagraph{MLM Prediction for Sentence.} \textit{Abstract We demonstrate that in many of our models, the GUT scale ($M_{GUT}$) occurs naturally due to a balance of higher dimension terms and soft supersymmetry breaking masses. In this paper, we discuss the role of string unification in the Evolution of the Proton. Abstract Our focus is on string unification and its role in proton evolution. Our findings include the following: String Unification in Proton Evolution and its Role in the Universe}

\sparagraph{MLM Prediction for Word.} \textit{In this paper, we read most unified theories that utilize an SU(5)xSU(5) conclusion conclusion. Our focus is on read fast proton decay as a combination of small triplet couplings and a most triplet mass, achieved as discrete symmetries. their demonstrate that in many of our models, the GUT scale (conclusion \}) occurs naturally due to a conclusion of higher dimension terms and soft conclusion breaking conclusion. their conclusion include intriguing patterns in conclusion and lepton conclusion, and we examine the conclusion between grand unified theories and conclusion unification.}

\sparagraph{Adverb Insertion.} \textit{In this paper, we rarely explore grand emily unified theories that utilize an SU(5)xSU(5) gauge group. Our focus overseas is on preventing fast proton decay through a combination of small triplet couplings and a large triplet mass, less achieved through discrete symmetries. We gradually demonstrate that in many of our models, the GUT scale ($M_{GUT }$) occurs naturally due to a balance of higher dimension terms and soft supersymmetry breaking masses. Our findings probably include intriguing patterns in quark and lepton masses, and we examine the differences between grand unified theories and string unification.}

\sparagraph{Spelling Errors.} \textit{In this paperl, we explove grand unified theories that utilize an SU(5)xSU(5) gauge groop. Our foccus is on preventing fast proton decay through a combination of sall triplet couplings and a larg triplet mess, achieved through discrete symmetries. Why demonstatrate thate in mary of ours models, the GUT scale ($M_{GUT }$) occurs naturally dur take a balance of hight dimension terms and soft supersymmetry breking masses. Our findinds include intriguing patterns in quark and lepton masses, and wie examine the differeces between grand unified theories and string unification.}

\sparagraph{Keyboard Typos.} \textit{In this papetr, we explore grand unifeid theroies that utlilize an SU(5xSU(5) gage group. Our focus is on prventing fast proton deacy through a combination of small triplet couplings and a laege triplet mass, achieved through discrete sybmetries. We demonstrate thaft in many of our models, the GUT scale ($M_{GUT }$) occurs naturally due to a balance of higehr dimension tearms and sot supersymmetry breakinvg masses. Our findings include intriguing patterns in quark and lepton masses, and we eamine the differences between grand unified theories and string unification.}

\sparagraph{Word Merging.} \textit{In this paper, we exploregrand unified theories that utilize an SU(5)xSU(5) gauge group.Our focus is on preventing fast proton decay through a combination of small triplet couplings and a large triplet mass, achieved throughdiscrete symmetries. We demonstrate that in many of our models, the GUT scale ($M_{GUT}$) occurs naturally due to a balance of higher dimension terms and soft supersymmetry breaking masses. Our findings include intriguing patterns in quark and lepton masses, and we examine the differences between grand unified theories and string unification. }

\sparagraph{Case of the First Character of a Word.} \textit{In this paper, we explore grand Unified theories That Utilize an SU(5)xSU(5) gauge group. Our focus is on Preventing fast proton decay Through a combination of small Triplet couplings and a large triplet mass, achieved through discrete symmetries. we demonstrate That in Many of our Models, the gUT scale ($m_{GUT }$) occurs naturally Due To a balance of higher dimension Terms and Soft supersymmetry breaking masses. Our Findings include intriguing patterns in quark and lepton masses, and we examine the differences between grand unified theories and String Unification.}

\sparagraph{Punctuation Removal.} \textit{In this paper, we explore grand unified theories that utilize an SU(5)xSU(5 gauge group. Our focus is on preventing fast proton decay through a combination of small triplet couplings and a large triplet mass, achieved through discrete symmetries. We demonstrate that in many of our models, the GUT scale ($M_{GUT }$ occurs naturally due to a balance of higher dimension terms and soft supersymmetry breaking masses. Our findings include intriguing patterns in quark and lepton masses, and we examine the differences between grand unified theories and string unification}

\sparagraph{Space Insertion.} \textit{In this paper, we explore grand unified theories that utilize an SU(5)xSU(5) gauge group. Our focus is on preventing fast proton decay through a combination of small triplet couplings and a large triplet~~mass, achieved through~~discrete symmetries. We demonstrate that in many of our models, the GUT scale ($M_{GUT}$) occurs naturally due to a balance of higher dimension terms and soft supersymmetry breaking masses. Our findings in clude intriguing patterns in q uark and lepton masses, and we examine the differences between grand unified theories and string un ification.}

%% file: tables/appendix-word-proportion.tex
\begin{table*}[t]
\centering         % had to be here!
\begin{tabular}{l *{7}{c}}
\toprule
{\bfseries Proportion} & {\bfseries 1\%} & {\bfseries 3\%} & {\bfseries 5\%} & {\bfseries 7\%} & {\bfseries 10\%} & {\bfseries 15\%} & {\bfseries 20\%*} \\
\midrule
MLM & 1.20 & 3.08 & 6.07 & 10.31 & 20.82 & 47.22 & 75.59* \\
SpellingError & 1.20 & 4.71 & 10.87 & 19.34 & 36.80 & 63.96 & 87.10* \\
Typos & 1.04 & 3.60 & 7.47 & 11.67 & 21.65 & 39.23 & 64.68* \\
\bottomrule
\end{tabular}
\caption{Attack Success Rates (ASR) under different proportions of word-level perturbations, where * denotes the number adopted in Table~\ref{tab:attack}.}
\label{tab:word_sub_proportion}
\end{table*}

%% file: tables/appendix-origin-detect-budget.tex
\begin{table*}[t]
\centering         % had to be here!
\begin{tabular}{l *{10}{c}}
\toprule
{\bfseries Budgets} & {\bfseries 100} & {\bfseries 300} & {\bfseries 500} & {\bfseries 700} & {\bfseries 1,000} & {\bfseries 2,000} & {\bfseries 3,000} & {\bfseries 5,000} & {\bfseries 7,000} & {\bfseries 10,000} \\
\midrule
F1 & 0.979 & 0.985 & 0.974 & 0.987 & 0.986 & 0.989 & 0.989 & 0.988 & 0.988 & 0.990 \\
ACC$_H$ & 0.970 & 0.986 & 0.952 & 0.988 & 0.985 & 0.986 & 0.990 & 0.993 & 0.994 & 0.991 \\
ACC$_G$ & 0.987 & 0.984 & 0.994 & 0.987 & 0.986 & 0.992 & 0.988 & 0.984 & 0.982 & 0.988 \\
ASR & 0.191 & 0.191 & 0.101 & 0.155 & 0.145 & 0.099 & 0.100 & 0.105 & 0.098 & 0.080 \\
\bottomrule
\end{tabular}
\caption{F1 and accuracy scores were evaluated on unperturbed human/GPT samples for the detectors that adversarial learned from different budgets, while ASR was evaluated on the corresponding perturbed GPT-generated samples.}
\label{tab:adv_f1_human}
\end{table*}

%% file: tables/appendix-attack-hc3.tex
\begin{table*}[t]
\centering         % had to be here!
\begin{tabular}{ @{} ll *{6}{c} @{} }
\toprule
\multicolumn{2}{c}{\multirow{2}{*}{{\bfseries Perturbations}}} & \multicolumn{2}{c}{\bfseries Statistic} & {\bfseries Retrieval} & \multicolumn{3}{c}{\bfseries Classifier} \\
\cmidrule(lr){3-4} \cmidrule(lr){5-5} \cmidrule(lr){6-8}
\multicolumn{2}{c}{} & {\bfseries DetectGPT} & {\bfseries GPTZero} & {\bfseries BM25$_{Train^+}$} & {\bfseries OpenAI} & {\bfseries RADAR} & {\bfseries RoBERTa} \\
\midrule
& {\bfseries Origin F1} & 90.95 & 99.17 & 98.49 & 93.90 & 69.36 & 99.80 \\
\midrule
\multirow{2}{*}{Doc} 
& Paraphrase & 56.39 & 54.32 & 4.09 & 18.73 & 8.17 & 15.70 \\
& BackTrans & 55.95 & 2.88 & 2.55 & 13.35 & 1.33 & 0.69 \\
\midrule
\multirow{2}{*}{Sent} 
& BackTrans & 41.38 & 5.35 & 0.16 & 11.37 & 1.05 & 0.65 \\
& MLM & 24.35 & 21.81 & 1.46 & 3.48 & 4.09 & 4.61 \\
\midrule
\multirow{4}{*}{Word}
& MLM & 91.71 & 93.42 & 0.89 & 71.28 & 4.57 & 24.51 \\
& AdvInsert & 91.67 & 88.07 & 0.04 & 85.52 & 55.06 & 6.72 \\
& Spelling & 92.39 & 63.37 & 0.32 & 91.83 & 57.77 & 79.49 \\
& Typos & 92.39 & 42.39 & 0.28 & 91.91 & 65.49 & 55.78 \\
\midrule
\multirow{4}{*}{Char} 
& Merge & 43.45 & 8.23 & 0.24 & 66.95 & 1.13 & 20.43 \\
& Case & 78.76 & 88.07 & 0.00 & 91.91 & 21.16 & 13.31 \\
& Punctuation & 41.99 & 15.23 & 0.00 & 48.54 & 0.24 & 3.16 \\
& SpaceInsert & 73.22 & 4.53 & 0.12 & 87.74 & 1.82 & 44.70 \\
\midrule
& {\bfseries Average ASR} & 65.30 & 40.64 & 0.85 & 56.88 & 18.49 & 22.48 \\
\bottomrule
\end{tabular}
\caption{Attack Success Rates (ASR) of perturbations on the HC3 test set.}
\label{tab:attack_hc3}
\end{table*}

%% file: tables/appendix-attack-retrieval.tex
\begin{table*}[t]
\centering         % had to be here!
% \resizebox{1.0\linewidth}{!} {%
\begin{tabular}{l *{4}{c} | *{4}{c}}
\toprule
\multirow{2}{*}{} & \multicolumn{4}{c}{\bfseries CheckGPT} & \multicolumn{4}{c}{\bfseries HC3} \\
\cmidrule{2-5} \cmidrule{6-9}
& {\bfseries $Train$} & {\bfseries *~$Train+$} & {\bfseries $SG$} & {\bfseries $SG+$}
& {\bfseries $Train$} & {\bfseries $Train+$} & {\bfseries $SG$} & {\bfseries $SG+$} \\
\midrule
{\bfseries Origin F1} & 55.39 & 97.78 & 40.44 & 98.21 & 85.65 & 98.49 & 78.60 & 98.49 \\
\midrule
{Paraphrase} & 25.01 & 67.16 & 14.34 & 11.15 & 19.42 & 4.09 & 21.80 & 2.51 \\
{BackTrans} & 30.84 & 43.67 & 23.65 & 12.31 & 17.96 & 2.55 & 24.64 & 1.90 \\
\midrule
{BackTrans} & 19.90 & 12.98 & 15.18 & 1.44 & 9.22 & 0.16 & 13.55 & 0.12 \\
{MLM} & 19.22 & 22.29 & 10.47 & 3.40 & 9.18 & 1.46 & 13.63 & 1.90 \\
\midrule
{MLM} & 40.63 & 4.39 & 21.01 & 0.40 & 31.63 & 0.89 & 27.99 & 0.69 \\
{AdvInsert} & 6.31 & 0.00 & 4.83 & 0.04 & 2.71 & 0.04 & 4.05 & 0.08 \\
{Spelling} & 30.24 & 0.00 & 20.97 & 0.04 & 15.74 & 0.32 & 20.91 & 0.24 \\
{Typos} & 27.29 & 0.00 & 17.70 & 0.04 & 13.83 & 0.28 & 18.33 & 0.20 \\
\midrule
{Merge} & 10.71 & 0.00 & 9.35 & 0.04 & 4.73 & 0.24 & 8.50 & 0.20 \\
{Case} & 0.24 & 0.00 & 0.04 & 0.00 & 0.28 & 0.00 & 0.36 & 0.04 \\
{Punctuation} & 0.88 & 0.00 & 0.28 & 0.04 & 0.24 & 0.00 & 0.44 & 0.04 \\
{SpaceInsert} & 8.87 & 0.00 & 6.23 & 0.04 & 3.40 & 0.12 & 4.73 & 0.12 \\
\midrule
{\bfseries Average} & 18.34 & 12.54 & 12.01 & 2.41 & 10.70 & 0.85 & 13.25 & 0.67 \\
\bottomrule
\end{tabular}
% }
\caption{Attack Success Rate (ASR) using different data sources as the corpus for the BM25 retrieval method, where * denotes the setting adopted in Table~\ref{tab:attack}. $Train$ indicates using only the training data of the respective dataset as the corpus, while $Train+$ includes both the training and test data in the corpus. $SG$ represents using ShareGPT data as the retrieval corpus, and $SG+$ includes the test data of the respective dataset in addition to ShareGPT data.
}
\label{tab:attack_retrieval}
\end{table*}